\documentclass[letterpaper,10pt,conference]{IEEEtran}

\IEEEoverridecommandlockouts  

\usepackage{xcolor}
\usepackage{hyperref}       
\usepackage{url}            
\usepackage{booktabs}       

\usepackage{amsfonts}       
\usepackage{nicefrac}       

\usepackage{graphicx}
\usepackage{multirow}
\usepackage{subfigure}

\let\labelindent\relax
\usepackage{enumitem} 

\usepackage{algorithm}
\usepackage{algpseudocode}

\usepackage{notation}
\usepackage{array}   
\newcolumntype{L}{>{$}l<{$}} 
\newcolumntype{C}{>{\centering\arraybackslash}p{2.4cm}}
\newcolumntype{D}{>{\centering\arraybackslash}p{1cm}}

\title{\LARGE \bf Model-based Reinforcement Learning from Signal Temporal Logic Specifications}

\author{Parv Kapoor$^{1}$, Anand Balakrishnan$^1$ and Jyotirmoy V. Deshmukh$^{1}$%
\thanks{$^{1}$%
Department of Computer Science,
University of Southern California,
Los Angeles, CA 90069
{\tt \small \{pk\_878,anandbal,jdeshmuk\}@usc.edu}}%
}

\renewcommand{\baselinestretch}{0.98}

\begin{document}

\maketitle
\thispagestyle{plain}
\pagestyle{plain}

\begin{abstract}

Techniques based on Reinforcement Learning (RL) are increasingly being used
to design control policies for robotic systems. RL fundamentally relies
on state-based reward functions to encode desired behavior of the robot and
bad reward functions are prone to exploitation by the learning agent, leading 
to behavior that is undesirable in the best case and critically dangerous 
in the worst. On the other hand, designing good reward functions for complex
tasks is a challenging problem. In this paper, we propose expressing desired
high-level robot behavior using a formal specification language known as 
Signal Temporal Logic (STL) as an alternative to reward/cost functions. We use
STL specifications in conjunction with model-based learning to design model 
predictive controllers that try to optimize the satisfaction of the STL 
specification over a finite time horizon. The proposed algorithm is empirically  
evaluated on simulations of robotic system such as a pick-and-place robotic 
arm, and adaptive cruise control for autonomous vehicles.

\end{abstract}

\section{Introduction}%
\label{sec:intro}
Reinforcement learning (RL)~\cite{sutton_reinforcement_2018} is a general class
of algorithms that enable automated controller synthesis, where a high-level
task/objective is coupled with repeated trial-and-error simulations to learn a
policy that satisfies the given specification.  Relatively recent developments
in deep learning have renewed interest in designing scalable RL algorithms for
highly complex
systems~\cite{mnih_humanlevel_2015,lillicrap_continuous_2019,schulman_highdimensional_2018,schulman_proximal_2017}.
While many recent works have focused on so-called \emph{model-free learning},
these algorithms are typically computationally expensive and model-free policies
are slow to train, as they are not \emph{sample efficient}. A promising
approach to increase sample efficiency is \emph{model-based reinforcement learning}
(MBRL)~\cite{atkeson_comparison_1997,kocijan_gaussian_2004,deisenroth_pilco_2011}.
The goal here is to (approximately) learn a predictive model of the system and
use this learned model to synthesize a controller using sampling-based methods
like model predictive control
(MPC)~\cite{camacho_model_2013,mayne_constrained_2000}. Another advantage of MBRL
is that as the learned dynamics models are task-independent they can be used to
synthesize controllers for various tasks in the same environment.


An important problem to address when designing and training reinforcement
learning agents is the design of \textit{reward
functions}~\cite{sutton_reinforcement_2018}.  Reward functions are a means to
incorporate knowledge of the goal in the training of an RL agent, using
hand-crafted and fine-tuned functions of the current state of
the system. Poorly designed reward functions can lead to the RL algorithm 
learning a policy that fails to accomplish the goal, cf.~\cite{leike_ai_2017}.  
Moreover, in safety-critical systems, the agent can learn a policy that
performs unsafe or unrealistic actions, even though it
maximizes the expected total reward~\cite{leike_ai_2017}. The problem of an RL
agent learning to maximize the total reward by exploiting the reward function,
and thus performing unwanted or unsafe behavior is called \textit{reward
hacking}~\cite{amodei_concrete_2016}.

With learning-enabled components being used increasingly commonly in real-world
applications like autonomous cars, there has been an increased interest in the
provably safe synthesis of controllers, especially in the context of
RL~\cite{fu_probably_2014,aksaray_qlearning_2016}. Work in safe RL has explored
techniques from control theory and formal methods to design provably safe 
controllers~\cite{fu_probably_2014,berkenkamp_safe_2017,wang_safe_2017}; and to
specify tasks by synthesizing reward functions from formal
specifications~\cite{aksaray_qlearning_2016,hasanbeig_logicallyconstrained_2018,balakrishnan_structured_2019}.

Reward hacking has been addressed by using \emph{Temporal Logics} like Linear
Temporal Logic (LTL) and Signal Temporal Logic (STL) to specify tasks and create
reward functions, as these logics can be used to specify complex tasks in a rich
and expressive manner. The authors in~\cite{aksaray_qlearning_2016,balakrishnan_structured_2019} 
explore the idea of using the robust satisfaction semantics of STL to define
reward functions for an RL procedure. Similar ideas were extended by to a 
related logic for RL-based control design for Markov Decision Processes (MDP)
in~\cite{li_reinforcement_2017}. 

Although these techniques are effective in theory, in practice, they do not
typically scale well to the complexity of the task and introduce large sampling
variance in learned policy. This is especially true in the case of tasks that
have a large planning horizons, or have sequential objectives. In contrast,
using Temporal Logic-based motion planning has been used successfully together
with model predictive control~\cite{farahani_robust_2015,sadigh_safe_2016},
where a predefined model is used to predict and evaluate future trajectories.

In this paper, we are interested in incorporating model-based reinforcement
learning with Signal Temporal Logic-based trajectory evaluation. The main
contributions of this paper are as follows:
\begin{enumerate}[leftmargin=1.5em,itemsep=1pt,topsep=1pt]
    \item We formulate a procedure to learn a deterministic predictive model of
    the system dynamics using deep neural networks. Given a state and a sequence
    of actions, such a predictive model produces a predicted trajectory over
    a user-specified time horizon.
    \item We use a cost function based on the quantitative semantics of STL to
    evaluate the optimality of the predicted trajectory, and use a black-box
    optimizer that uses evolutionary strategies to identify the optimal sequence 
    of actions (in an MPC setting). 
    \item We demonstrate the efficacy of our approach on a number of examples
    from the robotics and autonomous driving domains.
\end{enumerate}


\section{Preliminaries}%
\label{sec:prelims}
\subsection{Model-based Reinforcement Learning}%
\label{sub:mbrl}

In more traditional forms of controller design, like model predictive
control~\cite{camacho_model_2013}, a major step in the design process is to have
a model of the system dynamics or some approximation of it, which is in-turn
used as constraints in a receding horizon optimal control problem. In MBRL, 
this step is automated through a learning-based algorithm that approximates 
the system dynamics by assuming that the system can be represented by a Markov
Decision Process, and the transition dynamics can be learned by fitting a function
over a dataset of transitions.

A \emph{Markov Decision Process} (MDP) is a tuple \(M = \left(S, A, T,
\rho\right)\) where \(S\) is the state space of the system; \(A\) is the set of
actions that can be performed on the system; and \(T: S \times A \times S \to
[0,1]\) is the transition (or dynamics) function, where \(T(\bm{s}, \bm{a},
\bm{s}') = \Prob\of*{\bm{s}'\given \bm{s},\bm{a}}\); and \(\rho\) is a user-supplied
evaluation metric defined on either a transition of the system (like a reward
function) or over a series of transitions.


The goal of reinforcement learning (RL) is to learn a policy that maximizes 
\(\rho\) on the given MDP. 
In model-based reinforcement learning (MBRL), a model of the dynamics is used to
make predictions on the trajectory of the system, which is then used for action
selection. The forward dynamics of the model \(T\) is typically learned by
fitting a function \(\Model\) on a dataset \(\Dc\) of sample transitions
\((\bm{s},\bm{a},\bm{s}')\) collected from simulations or real-world
demonstrations. We will use \(\Model_\theta(\bm{s}_t,\bm{a}_t)\) to denote
the learned discrete-time dynamics function, parameterized by \(\theta\), that
takes the current state \(\bm{s}_t\) and an action \(\bm{a}_t\), and outputs a
distribution over the set of possible successor states \(\bm{s}_{t+\delta t}\).

\subsection{Model Predictive Control}%
\label{sub:mpc}

\emph{Model Predictive Control} (MPC)~\cite{camacho_model_2013} refers to a
class of online optimization algorithms used for designing robust controllers for
complex dynamical systems. MPC algorithms typically involve creating a model 
of the system (physics-based or data-driven), which is then used to predict 
the trajectory of a system over some finite horizon. The online optimizer in
an MPC algorithm computes a sequence of actions that optimizes the
predicted trajectories given a trajectory-based cost function (such as $\rho$) . This
optimization is typically done in a \emph{receding horizon} fashion, i.e., at
each step during the run of the controller, the horizon is displaced towards the
future by executing only the first control action and re-planning at every 
next step. Formally, the goal in MPC is to optimize $\rho$ over the space of all 
fixed length action sequences. For example, let an action sequence be denoted
\(\ActionSeq = (\bm{a}_t, \ldots, \bm{a}_{t+H-1})  \), then at every time step 
\(t\) during the execution of the controller for a finite \emph{planning horizon}
\(H\) we solve the following optimization problem:
\begin{align}\label{eq:mpc-optimization} \begin{split}
  \maximize_{\ActionSeq}\quad & \rho(\bm{\hat{s}}_t, \bm{a}_t,
  \bm{\hat{s}}_{t+1}, \ldots, \bm{a}_{t + H -1}, \bm{\hat{s}}_{t + H} ) \\
  \text{ where }\quad & \bm\hat{s}_t = \bm{s}_t\\ & \bm\hat{s}_{t + i + 1} =
  \Model(\bm{\hat{s}}_{t +i}, \bm{a}_{t+i}), \quad\text{ for } i \in 0,
\ldots, H -1 \end{split} \end{align}
where, \(\Model\) is the (approximate) system dynamics model.

Notice that the system dynamics are directly encoded as constraints in the
optimization problem: if the model is a linear system model, one can use Mixed
Integer Linear Programming, or Quadratic Programming to solve the optimization
problem at each step~\cite{kothare_robust_1996,richards_mixedinteger_2005}. For
nonlinear systems, this problem gets significantly harder, and can be solved
using nonlinear optimization techniques \cite{allgower2012nonlinear}, sampling 
based optimizers like Monte-Carlo methods~\cite{mesbah_stochastic_2016}, the 
Cross-Entropy Method~\cite{botev_crossentropy_2013}, or evolutionary strategies, 
like CMA-ES~\cite{hansen_cma_2006} and Natural Evolutionary Strategies~\cite{wierstra_natural_2014}.

\subsection{Signal Temporal Logic}%
\label{sub:signal-temporal-logic}

\textit{Signal Temporal Logic (STL)}~\cite{donze_robust_2010} is a real-time
logic, typically interpreted over signals over a dense-time domain that take
values in a continuous metric space (such as \(\Re^m\)). The basic primitive in
STL is a \emph{signal predicate} \(\mu\) -- a formula of the form
\(f(\bm{s}_t) \ge c\), where \(f\) is a function from the value domain
\(\domain\) to \(\Re\), and $c \in \Re$. STL formulas are then defined 
recursively using Boolean combinations of sub-formulas, or by applying an 
interval-restricted temporal operators to a sub-formula. In this paper, we 
consider a restricted fragment of STL (where all temporal operators are
unbounded). The syntax of our fragment of STL is 
formally defined as follows:
\begin{equation} \label{eq:stl_syntax}
  \varphi ::= \mu
              \mid \neg\varphi
              \mid \varphi \wedge \varphi
              \mid \alw\varphi
              \mid \ev \varphi
              \mid \varphi \until \varphi
\end{equation}

Here, \(I = [a,b]\) denotes an arbitrary time-interval, where \(a,b\in\domain\).
The Boolean satisfaction semantics of STL are defined recursively over the 
structure of an STL formula; in lieu of the Boolean semantics, we instead 
present the {\em quantitative semantics} of STL and explain how these 
relate to the Boolean semantics. The quantitative semantics of an STL formula 
$\varphi$ are defined in terms of a {\em robustness} value $\rho(\varphi,\signal,t)$
that maps the suffix of the signal $\signal$ starting from time $t$ to a 
real value. This value (approximately) measures a signed distance by which the 
signal can be perturbed before its (Boolean) satisfaction value changes. A positive 
value implies that the signal is satisfied, while a negative robustness value
implies that the systems violates the specification. Formally, the approximate
robustness value (or simply robustness) is defined using the following recursive
semantics \cite{donze_robust_2010}:
\begingroup
\def\arraystretch{1.7}
\begin{equation}
    \begin{array}{r@{\hspace{.2em}}c@{\hspace{.4em}}l}
    \rho(f(\signal) \ge c,  \signal, t) & \equiv & f(\signal_t) -c \\
    \rho(\neg \varphi, \signal, t) & \equiv & -\rho(\varphi,\signal,t) \\
    \rho(\varphi_1 \wedge \varphi_2, \signal, t) & \equiv &
            \min(\rho(\varphi_1,\signal,t),\rho(\varphi_2,\signal,t)) \\
    \rho(\alw \varphi,  \signal, t) & \equiv & 
        \inf_{t'\ge t} \rho(\varphi,\signal,t') \\
    \rho(\ev \varphi,  \signal, t) & \equiv & 
        \sup_{t'\ge t} \rho(\varphi,\signal,t') \\
    \rho(\varphi_1 \until \varphi_2,  \signal, t) & \equiv & 
        \displaystyle\sup_{t' \ge t}\!\! \min\!\left(\!\!
        \begin{array}{l}
            \rho(\varphi_2, \signal, t'), \\
            \displaystyle\inf_{t'' \in [t,t')} \rho(\signal,\varphi_1,t'')
        \end{array}\!\!\right)\!\!
    \end{array}
\end{equation}
\endgroup
\noindent The convention is that $\signal \models \varphi$ if $\rho(\varphi,\signal,0) \ge 
0$, and $\signal$ does not satisfy $\varphi$ otherwise. 

%
%
%
%
%
%


\section{Robust Controller Synthesis with Approximate Model}%
\label{sec:controller-synthesis}

We now present a framework that combines the use of model-based reinforcement
learning (MBRL) and sampling-based model predictive control (MPC) to maximize
the robustness value of a trajectory against a given STL formula. We first learn
a deterministic neural network model of the system, \(\Model\), which can be
used to predict what the next state will be, given the current state and an
action. We then use this model in an MPC setting described
in~\autoref{eq:mpc-optimization}, where we use the model to sample trajectories
over a finite horizon, \(H\), and use CMA-ES to optimize a sequence of actions,
\(\bm{A}_t^{(H)}\), that maximizes the robustness of the predicted trajectory
against an STL specification, \(\varphi\).

\subsection{Learning the System Dynamics}%
\label{sub:learning_the_system_dynamics}

\begin{algorithm}
  \caption{\label{alg:model_training}Learning the system dynamics model.}
  \begin{algorithmic}[1]
    \State Initialize empty dataset, \(\Dc\)
    \For{\(i \in 1,\ldots, N_\text{traj}\)}
      \For{Each time step \(t\)}
      \State \(\bm{a}_t \sim \textsf{Uniform}(\cdot), \bm{s}_{t+1} \sim \textsf{Env}(\bm{s}_t,\bm{a}_t)\)
      \State \(\Dc \gets (\bm{s}_t, \bm{a}_t, \bm{s}_{t+1})\)
      \EndFor{}
    \EndFor{}
    \State \(\theta \gets \textsf{SGD}(\Dc)\)
  \end{algorithmic}
\end{algorithm}

In MBRL, an approximate model of the system dynamics is learned through repeated
sampling of the environment. This model can be represented using various means,
including by the use of Gaussian
Processes~\cite{deisenroth_pilco_2011,deisenroth_gaussian_2015} and, more
recently, Deep Neural
Networks~\cite{nagabandi_neural_2017,chua_deep_2018,nagabandi_deep_2019}. Here,
we represent the learned dynamics, \(\Model_\theta(\bm{s}_t, \bm{a}_t)\) as a
deep neural network, where \(\theta\) is the vector of parameters of the NN.\@
The learned dynamics, \(\Model_\theta\), is \emph{deterministic}, i.e., it
outputs a single prediction on the what the change in the state of the system,
\(\Delta \bm{\hat{s}}_t = \bm{s}_{t+1} - \bm{s}_t\) will be,  rather than
outputting a distribution on the predicted states.\footnote{We pick a
deterministic model as, in our chosen case studies, we notice that a lot of system noise is not
time-varying, and thus, it suffices to predict the \emph{mean} predicted state
for some state-action pair.} The same design decisions were taken
in~\cite{nagabandi_neural_2017}, where they use MBRL as a preceding step to
accelerate the convergence of a model-free reinforcement learning algorithm. The
steps to train the model can be seen in~\autoref{alg:model_training}.

\subsubsection*{Collecting data}%

To generate the dataset, \(\Dc\), of samples \(\left( \bm{s}, \bm{a}, \bm{s}'
\right)\), we assume that we have access to some simulator, for example, the
OpenAI Gym~\cite{brockman_openai_2016}. We sample a large number of trajectories
(and hence, a large number of transitions), where each trajectory start at some
initial state initial state \(\bm{s}_0\) from the environment, and actions are
sampled from a \emph{random controller}, i.e., we pick actions from a uniform
distribution over the action space of the environment.


\subsubsection*{Preprocessing the dataset}%

After generating the data, we calculate the target as \(\Delta \bm{\hat{s}} =
\left(\bm{s'} - \bm{s} \right)\). We then calculate the mean and standard
deviation for the inputs \(\left( \bm{s}, \bm{a} \right)\) and target outputs.
We rescale the data by subtracting the mean from it and dividing it by standard
deviation. This normalization aids the model training process and leads to
accurate predictions for all dimensions in the state space.

\subsubsection*{Training the model}%

We train \(\Model_\theta(\bm{s}, \bm{a})\) by minimizing the sum of squared
error loss for each transition, \((\bm{s}, \bm{a}, \bm{s}')\), in the dataset as
follows:
\begin{equation}\label{eq:mse-loss}
  \Lc(\theta) = \sum_{(\bm{s}, \bm{a}, \bm{s'}) \in \Dc_\text{train}} {\left( \bm{s}' - \Model_\theta(\bm{s}, \bm{a}) \right)}^2,
\end{equation}
where \(\Dc_\text{train}\) is a partition on the dataset \(\Dc\) allocated for
training data. We use the remaining partition, \(\Dc_\text{val}\), of the
dataset as a validation dataset, over which we calculate the same loss described
in~\autoref{eq:mse-loss}. 
The loss minimization is carried out using stochastic gradient descent, where
the dataset is split into randomly sampled batches and the loss is minimized
over these batches.

\subsection{Sampling-based Model Predictive Control}%
\label{sub:sampling_based_model_predictive_control}

\begin{algorithm}
  \caption{\label{alg:sample-mpc}Model Predictive Control using \(\Model_\theta\).}
  \begin{algorithmic}[1]
    \For{Each time step \(t\) in episode}
      \For{ \(1 \ldots N_\text{iter}\) and \(1 \ldots N_\text{samples}\) }
        \State \(\bm{A}_t^{(H)} = (\bm{a}_t, \ldots, \bm{a}_{t+H-1}) \sim \text{CMA-ES}()\)
        \For{\(i \in 0, \ldots, H-1\)}
          \State \(\bm{\hat{s}}_{t + i + 1} = \Model_\theta(\bm{\hat{s}}_{t+i},\bm{\hat{a}}_{t+i})\)
        \EndFor{}
        \State Compute cost \(\rho(\bm{\hat{s}}_t, \bm{a}_t, \bm{\hat{s}}_{t+1}, \ldots, \bm{a}_{t + H -1}, \bm{\hat{s}}_{t + H} )\)
        \State Update \(\text{CMA-ES}()\)
      \EndFor{}
      \State Execute first action \(\bm{a}^*\) from optimal sequence \(\bm{A}_t^{(H)} \sim \text{CMA-ES}()\)
    \EndFor{}
  \end{algorithmic}
\end{algorithm}

Once we have an approximate dynamics model, \(\Model_\theta\), where the
parameters \(\theta\) have been trained as in the previous section, we can use
this model in a model predictive control setting.

At state \(\bm{s}_t\), our goal is to compute a sequence of actions,
\(\ActionSeq = (\bm{a}_t, \ldots, \bm{a}_{t+H-1})\), for some finite horizon,
\(H\). As stated earlier, this can be formulated as an optimization problem over
sampled trajectories, as described in~\autoref{eq:mpc-optimization}, and the
optimization is performed using the CMA-ES~\cite{hansen_cma_2006} black-box
optimizer. During each iteration of the optimizer, a number of action sequences
are sampled from the optimizer's internal distribution, and a trajectory is
computed for each action sequence by repeatedly applying
\(\Model(\bm{\hat{s}}_{t + i},\bm{a}_{t + i})\) to get \(\bm{\hat{s}}_{t+i+1}\)
(here, \(\bm{\hat{s}}_t = \bm{s}_t\)). These trajectories are then used as an
input to compute the robust satisfaction value, or robustness, of the sequence
of actions, given a STL specification \(\varphi\), which is used as the
objective function to maximize by CMA-ES.\@ At each step, we execute only the
first action given to us by CMA-ES, and perform the optimization loop at each
step. The pseudocode for this can be seen in~\autoref{alg:sample-mpc}.

\section{Experimental Results and Discussions}%
\label{sec:experiments}
\begin{figure}
    \centerline{%
    \subfigure[Fetch Robot]{%
    \includegraphics[width=0.20\textwidth]{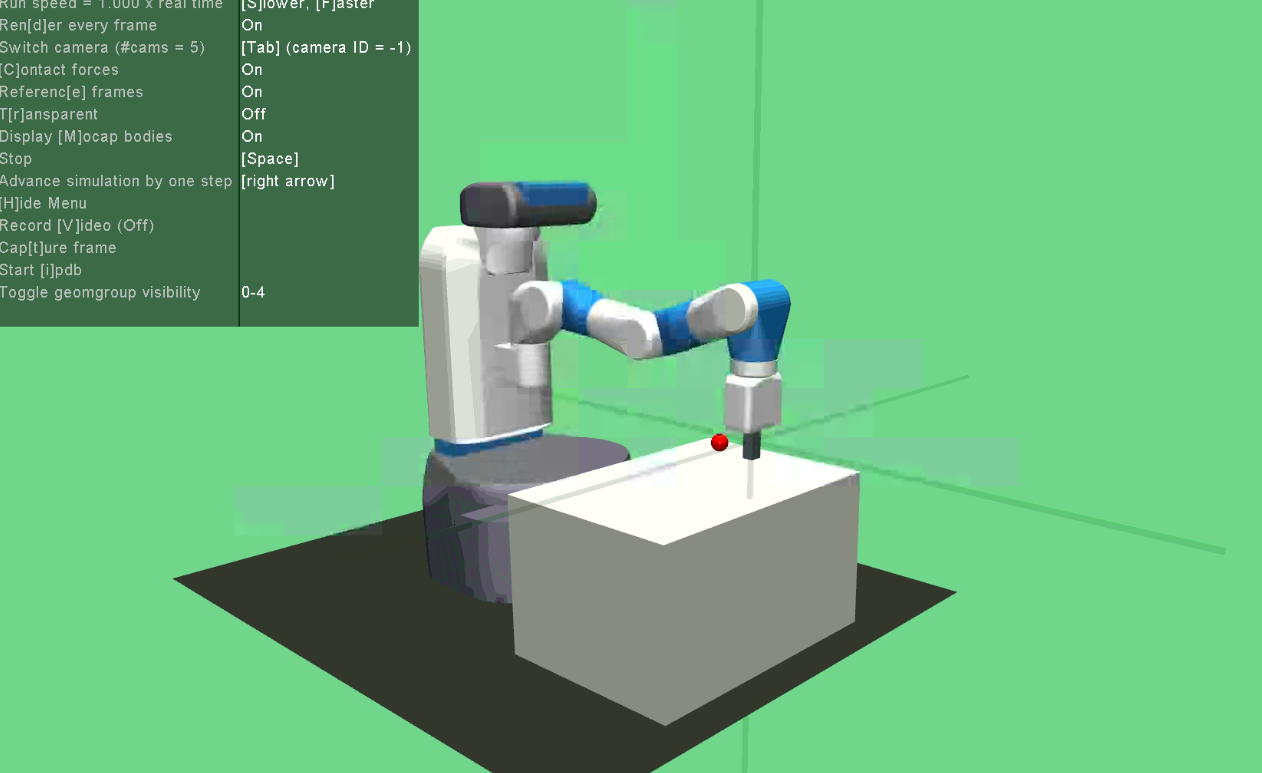}}
    \hfil%
    \subfigure[Highway Cruise Control]{%
    \includegraphics[width=0.20\textwidth]{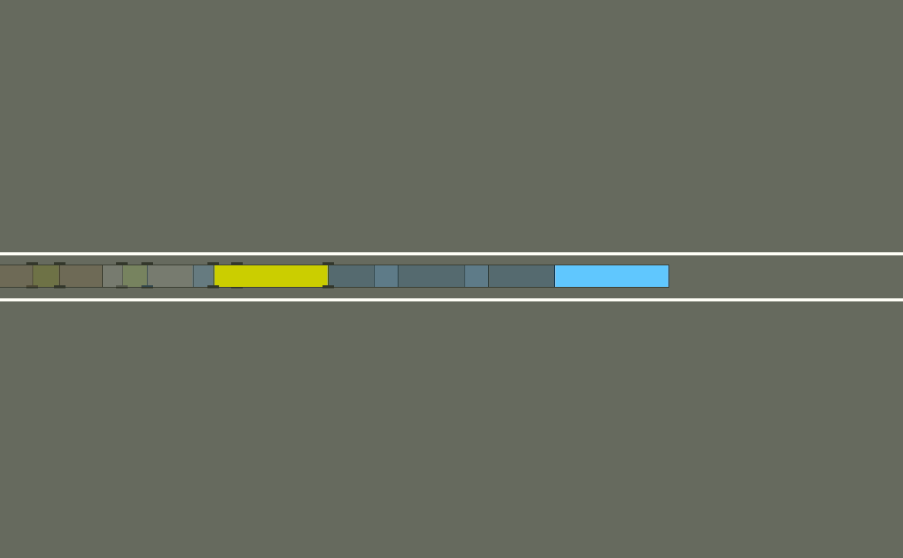}}%
    }
    \caption{Examples of environments used for evaluation.\label{fig:environment_pics}}
\end{figure}

\begin{figure*}
    \centerline{%
    \subfigure[Cartpole]{%
    \includegraphics[width=0.20\textwidth]{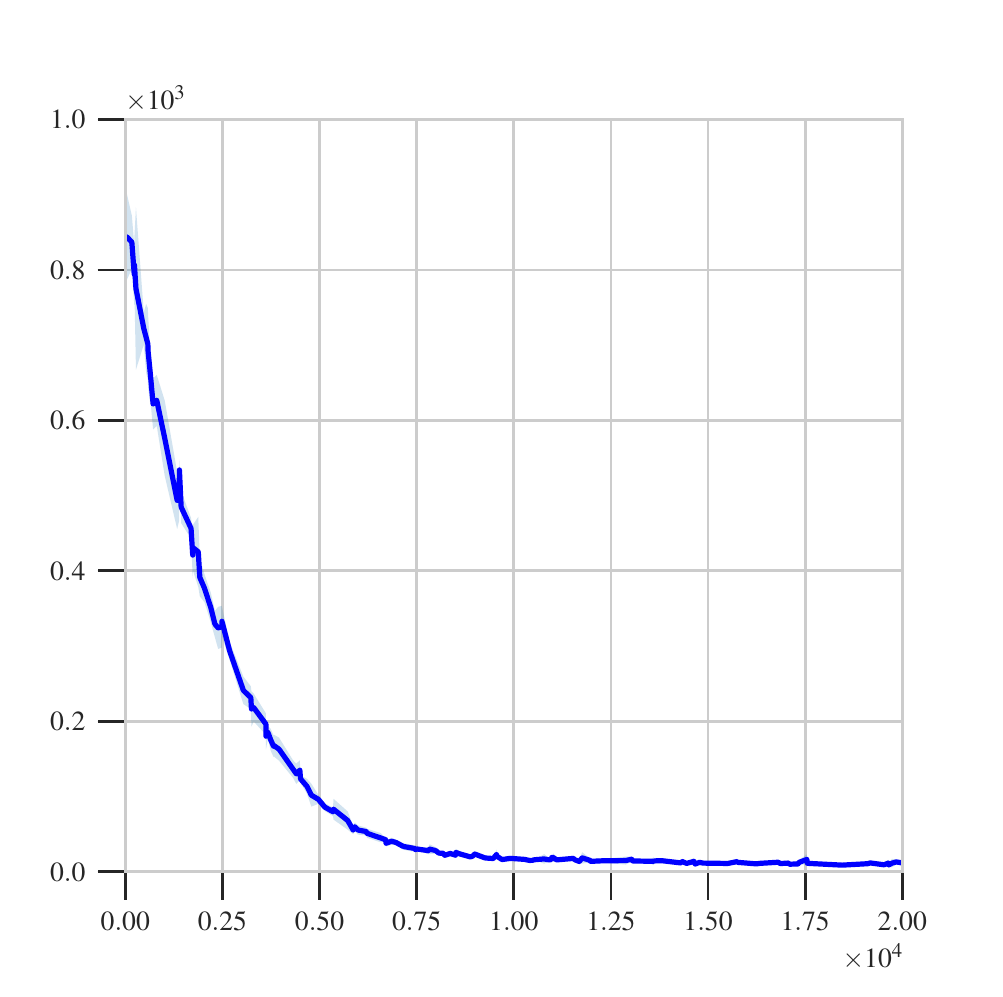}}
    \hfil%
    \subfigure[Mountain Car]{%
    \includegraphics[width=0.20\textwidth]{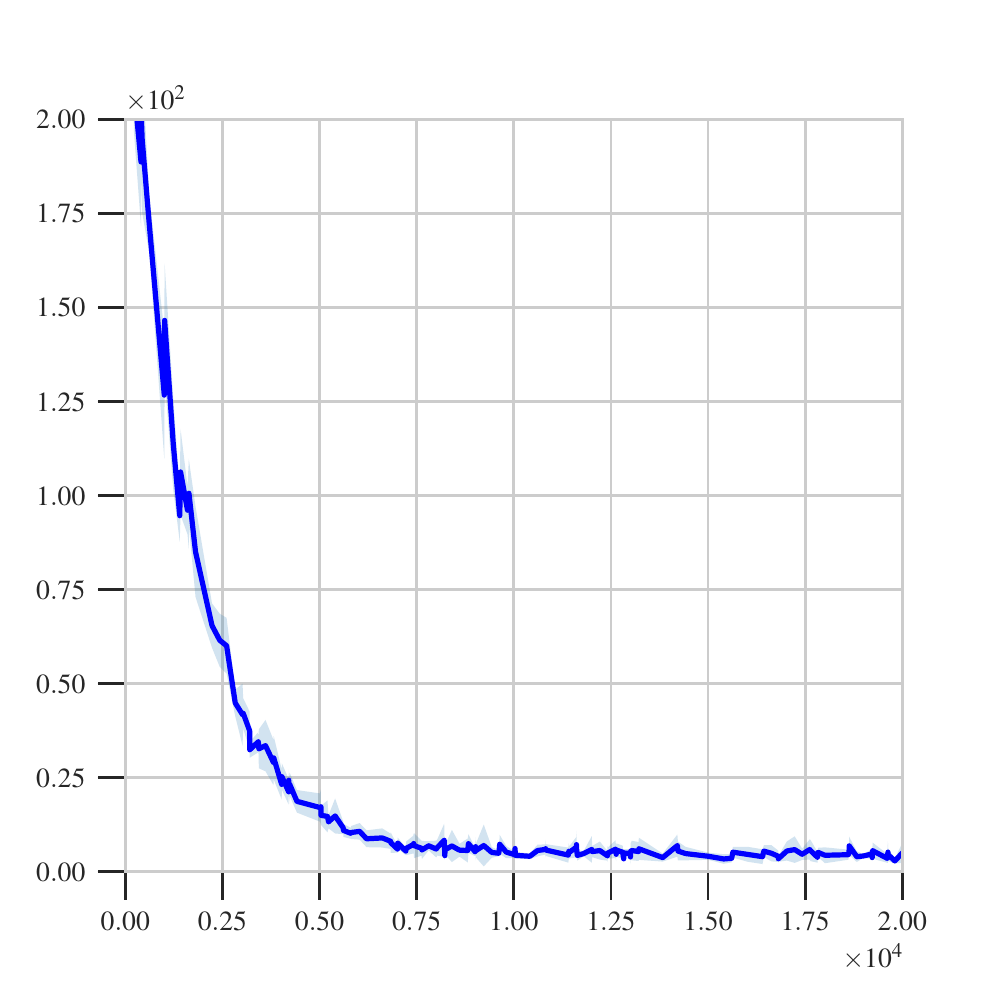}}%
    \hfil%
    \subfigure[Highway]{%
    \includegraphics[width=0.20\textwidth]{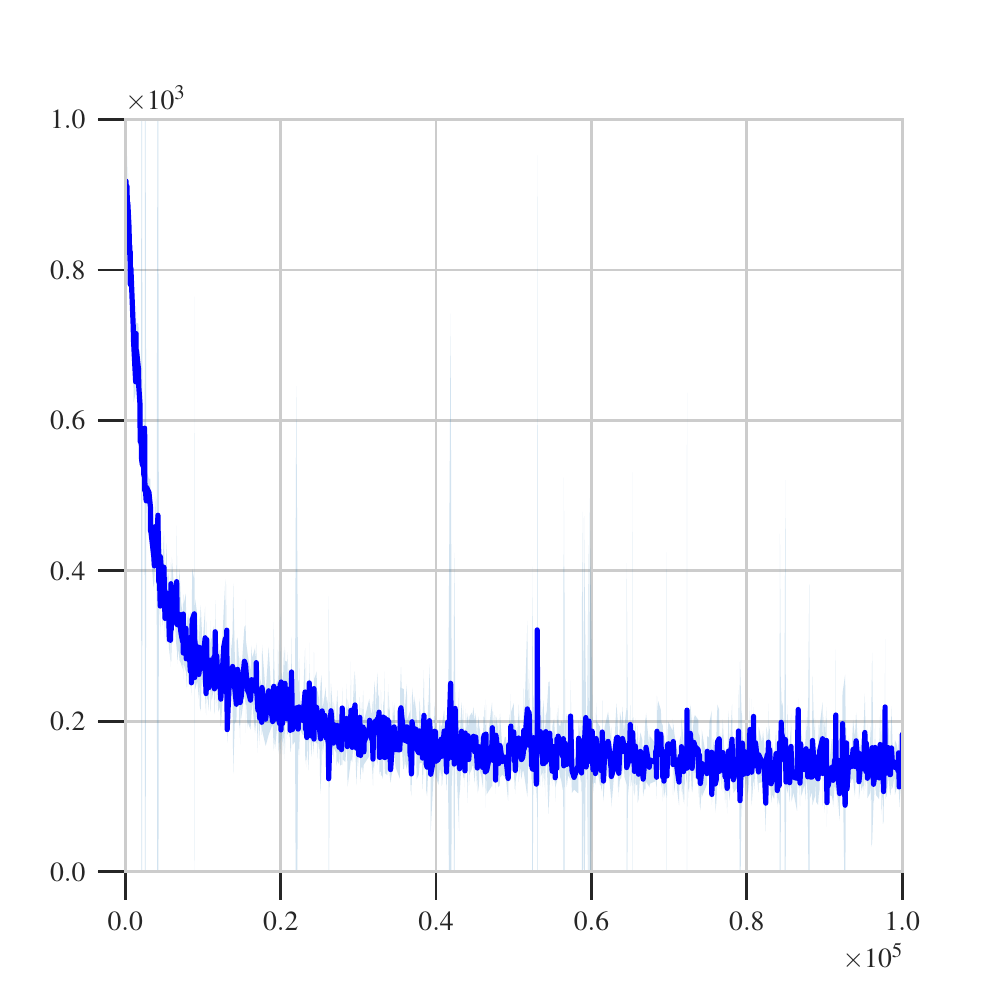}}%
    \hfil%
    \subfigure[Parking Lot]{%
    \includegraphics[width=0.20\textwidth]{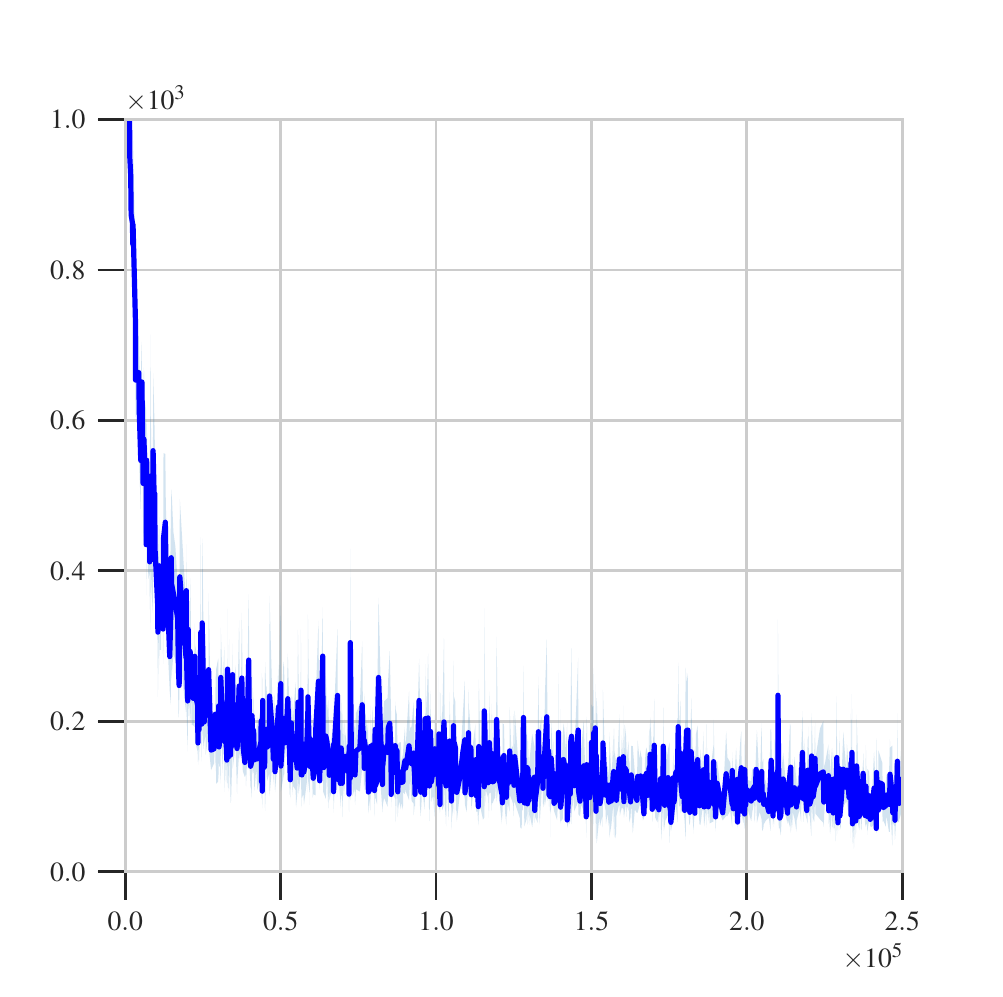}}%
    \hfil%
    \subfigure[Fetch robot]{%
    \includegraphics[width=0.20\textwidth]{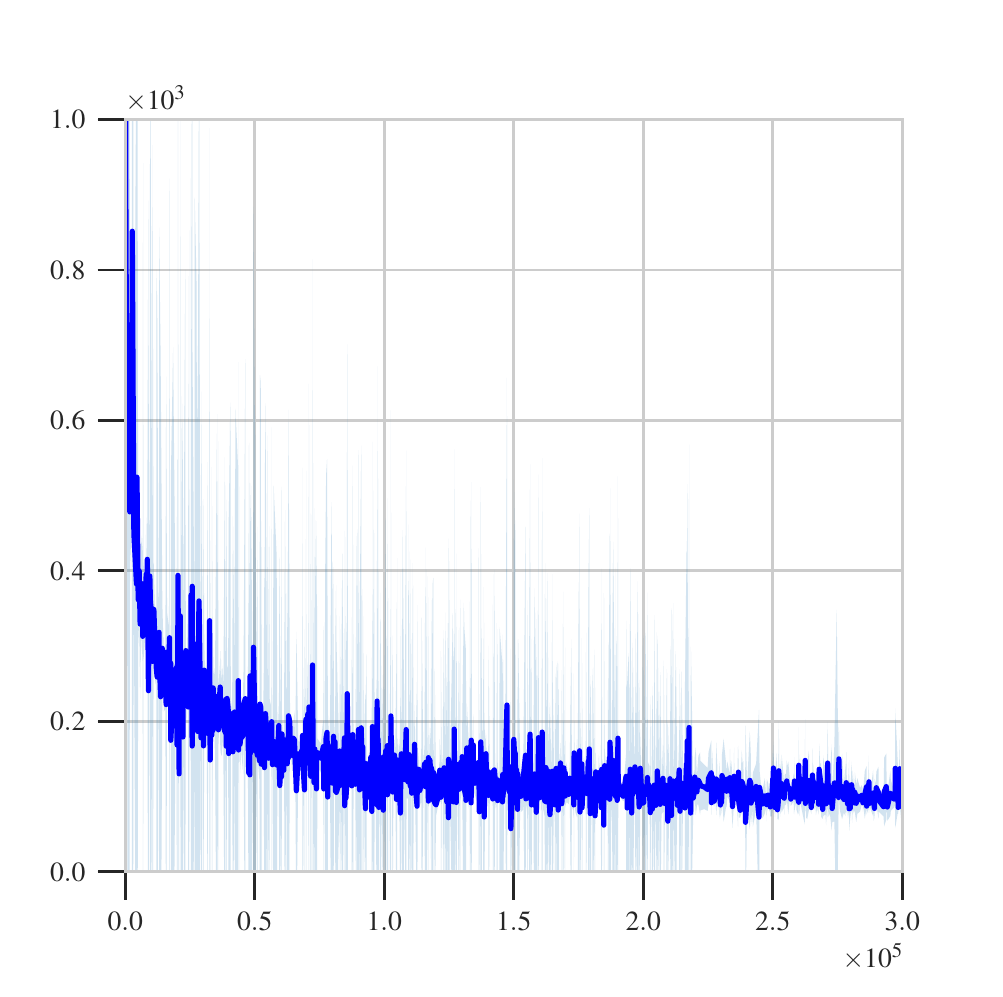}}%
    }
    \caption{Model learning losses for each environment. The plot measures the
    average decay of loss with the number of training iterations across 7 different training runs.\label{fig:losses}}
\end{figure*}

We evaluate our framework, we train dynamics models for multiple different
environments, and evaluate the model performance in the model predictive control
setting. In this section, we will describe the environments we used, the tasks
defined on them, and the results of using the proposed deep model predictive
control framework on them, which can be seen in~\autoref{tab:results}.

\textit{Remark}. It should be noted that to train the models, we chose a random
controller to create the transition dataset as, in the environments we are
studying, we saw that a random controller strategy for exploration of the state
and action spaces works either just as good as, or sometimes better than, other
methods that trade off between exploration for new transitions and exploitation
of previous knowledge.

\subsection{Cartpole}%
\label{sub:cartpole}

This is the environment described in~\cite{barto_neuronlike_1983}, where a pole
is attached to a cart on an unactuated joint. The cart moves on a frictionless
track, and is controlled by applying a force to push it left or right on the
track. The state vector for the environment is \(\left(\theta, x, \dot{\theta},
\dot{x} \right)\), where \(\theta\) is the angular displacement of the pole from
vertical; \(x\) is the displacement of the cart from the center of the track;
and \(\dot{\theta}\) and \(\dot{x}\) are their corresponding velocities.

The requirement, as described in~\cite{barto_neuronlike_1983}, is to balance the
pole for as long as possible on the cart (maintain angular displacement between
\(\pm 12^\circ\)). For our experiments, we try to maximize the robustness of the
following specification:
\begin{equation}
  \varphi := \always\left(\abs{x} < 0.1 \land \abs{\theta} < 12^\circ\right)
\end{equation}
The above specification requires that, at all time steps, the cart displacement
shouldn't exceed a magnitude of \(0.1\) and the angular displacement of the pole
should not exceed a magnitude of \(12^\circ\).

In~\autoref{tab:results}, we see that for a horizon length of \(10\), we are
able to maximize the robustness --- which is upper bounded by \(0.1\) as, when
the positional and angular displacement are \(0\), the robustness will be the
minimum distance from violation, which is \(0.1\). The minor loss of robustness
can be attributed to the stochasticity in the environment dynamics, making it
impossible to perfectly balance the pole from time \(0\). Moreover, across multiple evaluation runs, the
controller is also able to maximize the rewards obtained from the environment
under the standard rewarding scheme, which is \(+1\) for each time step that the
pole is balanced on the cart, for a maximum of 200 time steps. Specifically, the controller is able to attain a total reward of 200 consistently, across multiple runs.

\subsection{Mountain Car}%
\label{sub:mountain_car}

This environment describes a mountain climbing task, as described
in~\cite{moore_efficient_1990}. The environment consists of a car stuck at the
bottom of a valley between two mountains, and the car can be controlled by
applying a force pushing it left or right. The state vector for the cart is on a
one-dimensional left-right track, \(x, \dot{x}\), where \(x\) is the
displacement of the cart on the track, and \(\dot{x}\) is the car's velocity
along the horizontal track.

The goal of this environment is to synthesize a controller that pushes the car
to the top of the mountain on the right. Under a traditional rewarding scheme,
the controller is rewarded a large positive amount for reaching the top of the
mountain, and a small negative amount for every step it doesn't reach the top.
We encode this specification using the following STL formula, which requires
that the cart should \emph{eventually} reach the top of the right side hill:
\begin{equation}
  \varphi := \eventually\left(x > 0.4\right),
\end{equation}
where \(0.4\) is the horizontal coordinate of the top of the mountain on the
right side.

Here, since the goal is to \emph{reach} a configuration, the maximum robustness
is positive only when it actually reaches the goal. Thus, any positive
robustness implies a successful controller, as seen in~\autoref{tab:results}.
The classical rewarding scheme is as follows,
for every time step that the car hasn't reached the goal, it gets a reward of
\(-0.1*action^{2}\), and only gets a positive \(100\) reward when it actually reaches the top of the hill. A solution algorithm is considered to "solve" the problem if it achieves a reward of 90.0 averaged over multiple evaluations. One key take away from this is that our controller is able to achieve this reward despite not having this requirement encoded in specifications explicitly!

\subsection{Fetch robot}%
\label{sub:fetch_robot}

We use the OpenAI Gym \emph{FetchReach}
environment~\cite{plappert_multigoal_2018}, where the goal of the environment is
design a controller to move the arm of a simulated Fetch manipulator
robot~\cite{wise_fetch_2016} to a region in 3D space. The state vector of the
environment is a 16 dimensional vector containing the position and orientation
of the robots joints and end-effector. The below specification is defined solely
on the position of the end-effector, denoted \(\left(x, y, z\right)\), and the
position of the goal \( \left( x_g, y_g, z_g \right)\):
\begin{equation}
  \varphi := \eventually\left(\abs{x_g - x} < 0.1 \land \abs{y_g - y} < 0.1 \land \abs{z_g - z} < 0.1\right)
\end{equation}

Similar to the previous \emph{reach} specification, we see that the controller
learned for the \emph{FetchReach} environment is able to attain a positive
reward on average and a 100 percent success rate, thus satisfying the specification.

\subsection{Adaptive Cruise-Control}%
\label{sub:adaptive_cruise_control}

Here, we simulate an adaptive cruise-control (ACC) scenario on a highway
simulator~\cite{leurent_environment_2018}. Here, the goal is to synthesize a
controller for a car (referred to as the \emph{ego} car) that safely does cruise
control in the presence of an adversarial (or \emph{ado}) car ahead of it. The
environment itself is a single lane, and the \emph{ego} car is controlled only
by the throttle/acceleration applied to the car. The \emph{ado} agent ahead of
the car accelerates and decelerates in an erratic manner, in an attempt to cause
a crash.

The state space of the environment is \( \left( x_\text{ego}, v_\text{ego},
a_\text{ego}, d_\text{rel}, v_\text{rel} \right)\), where
\(x_\text{ego}\), \(v_\text{ego}\), and \(a_\text{ego}\) are the position,
velocity, and acceleration of the \emph{ego} car on the highway; and
\(d_\text{rel}\) and \(v_\text{rel}\) are, respectively, the relative
displacement and velocity of the ado car relative to the \emph{ego} car's 
coordinate frame. We have the following requirements for the cruise controller:
\begin{equation}
  \varphi := \eventually\always( 50 > d_\text{rel} > 15)
\end{equation}

To aid the model learning process, augment the above state space with the
relative acceleration of the \emph{ado} car, \(a_\text{rel}\). This is done by
maintaining a difference between \(v_\text{rel}\) between two consecutive time
steps. Augmenting the state space with this manual (and trivial) estimation
allows for more stable and consistent predictions by the model.

The above specification is a \emph{stable} goal specification, that is, the
formula requires that the \emph{ego} car comes so some stable configuration
eventually in the future. Thus, for any final configuration that maintains the
relative distance between the \emph{ego} and \emph{ado} cars at a value between
\(15\) and \(50\) satisfies the specification. In~\autoref{tab:results}, we see
that the robustness \(\rho\) is consistently positive and, since the robustness
is the distance to violation, we see that it is well within the bounds (by
\(\sim 7\) units).

\subsection{Parking Lot}%
\label{sub:parking_lot}

We use the same simulator~\cite{leurent_environment_2018} used in the previous
environment to simulate a parking lot, where we design a controller to park a
car in a specific spot. Here, the state of the system is described on the 2D
coordinate space of the parking lot surface, \[\left( x, y, v_x, v_y,
\cos(\theta), \sin(\theta), x_g, y_g \right),\] where \(x, y\) are the position;
\(v_x, v_y\) are the velocity components; \(\theta\) is the bearing angle of the
car; and \(x_g, y_g\) are the coordinates of the parking spot.

We evaluate the controller using the following specifications:
\begin{equation}
  \varphi := \eventually\left(\abs{x_g - x} < 0.02 \land \abs{y_g - y} < 0.02\right)
\end{equation}

Similar to \emph{Mountain Car} and \emph{FetchReach}, we see that this
specification is also a \emph{reach} requirement. Thus, any positive value close
to \(0\) is considered a success. This is exactly what we see
in~\autoref{tab:results}.

\subsection{Discussion and Analysis}

\begin{table*}
\centering
\caption{Hyperparameters and Results.}%
\label{tab:results}
\begin{minipage}{\linewidth}
\begin{tabular}{lCCCDCC}
\toprule
\textbf{Environment} & \multicolumn{4}{c}{\bf Hyperparameters} & \multicolumn{2}{c}{\textbf{ Results} (averaged across 30 runs)} \\
                \cmidrule(lr){2-5}
                \cmidrule(lr){6-7}
            & No.\ of training trajectories & No.\ of optimizer iterations & No.\ of optim.\ samples per iteration & Horizon & Trajectory Robustness & Vanilla rewards\footnote{Not used for training. It is used purely as a means to evaluate how well the controller that optimizes the STL robustness does in maximizing the original reward function defined for the environment.} \\
              & \(N_\text{traj}\) & \(N_\text{iter}\) & \(N_\text{samples}\) & \(H\)  & \(\rho\)                                         & \\ 
\toprule
Cartpole      & 2000          & 5                 & 1000                 & 10          & \(0.069 \pm \expnumber{8}{-3}\)                  & \(200.0 \pm 0\) \\ 
Mountain Car  & 2000          & 2                 & 1000                 & 50          & \(0.047 \pm \expnumber{1}{-3}\)                  & \(89.06 \pm 4.57\) \\
Fetch         & 2000         & 7                 & 500                  & 10          & \( 0.067\pm \expnumber{6}{-3}\)                  & -- \\
ACC           & 400          & 7                 & 500                  & 2           & \(7.679 \pm 5.47\)                               & -- \\
Parking Lot   & 400          & 5                 & 5                    & 5          & \(\expnumber{1.69}{-2} \pm \expnumber{3.8}{-3}\) & -- \\

\bottomrule
\end{tabular}
\end{minipage}
\end{table*}

In the experimental results in~\autoref{tab:results}, we see that using a
deterministic model suffices for predicting trajectories as we are able to
maximize the robustness and, if applicable, the rewards for a given system and
specification. This even applies to complex environments like \emph{Adaptive
Cruise-Control}, where we have an adversarial car performing erratic behavior.
We speculate that this is due to the noise in the system being time-invariant,
thus predicting the mean configuration of the system is sufficiently accurate
for the controller.


\section{Related Work}%
\label{sec:related-work}
\subsubsection*{Model-based Reinforcement Learning}%
\label{sub:mbrl-related-work}

Model-free deep reinforcement learning algorithms based on
Q-learning~\cite{mnih_humanlevel_2015,gu_continuous_2016}, actor-critic
methods~\cite{lillicrap_continuous_2019,schulman_highdimensional_2018,mnih_asynchronous_2016},
and policy gradients~\cite{silver_deterministic_2014,schulman_proximal_2017} have been
used to directly learn policies for specific tasks in highly complex systems including
simulated robotic locomotion, driving, video game playing, and navigation. While these
methods have been successful and popular, they tend to require an incredibly large
number of samples and extensive computational capabilities.  In contrast, in MBRL, the
goal is to learn a model of the system through repeatedly acting on the environment and
observing state transition dynamics, and then use this model to synthesize a
controller~\cite{atkeson_comparison_1997} or use an under-approximation to aid in
exploration of the state space~\cite{thrun_efficient_1992}.

Gaussian Processes~\cite{rasmussen_gaussian_2006} have been used extensively as
non-parametric Bayesian models in various MBRL problems where data-efficiency is
critical~\cite{kocijan_gaussian_2004,nguyen-tuong_local_2009,deisenroth_pilco_2011,deisenroth_gaussian_2015}.
In recent years, neural network-based models have generally supplanted GPs in
MBRL~\cite{gal_improving_2016,fu_oneshot_2016,nagabandi_neural_2017,depeweg_decomposition_2018,chua_deep_2018,nagabandi_deep_2019},
as neural networks have fixed-time runtime execution, and can generalize better to
complex, non-linear, high-dimensional systems. Moreover, they perform much better than
Gaussian Processes in the presence of large amount of data, as in the case of RL.\@

Recent work has used sampling-based model predictive control (MPC) along with deep neural network dynamics models, where an optimizer outputs a sequence of actions which are used to predict trajectories, and, in-turn, the cost of the sequence of actions. The work by~\cite{nagabandi_neural_2017} uses a deterministic NN model with a random search optimizer for computing an action sequence. On the other hand, the work in~\cite{chua_deep_2018} uses an \emph{ensemble of probabilistic models} to output a probability distribution over the future states, and uses the Cross-Entropy Method for optimizing the sequence of actions that maximize predicted trajectory rewards.


\subsubsection*{Formal Methods in Controller Synthesis}%
\label{sub:safe-cps-related-work}

Temporal logics have been used extensively in the context of cyber-physical systems to
encode temporal dependencies in the state of a system in an expressive
manner~\cite{donze_automotive_2015,jin_mining_2015,bartocci_specificationbased_2018}.
Originating in program synthesis and model checking literature, temporal logics like
Linear Temporal Logic, Metric Temporal Logic, and Signal Temporal Logic, have been used
extensively to synthesize controllers for complex tasks, including motion
planning~\cite{lahijanian_motion_2010,garg_controllyapunov_2019} and synthesizing
controllers for various robot systems, including
swarms~\cite{moarref_automated_2020,sadigh_safe_2016,lindemann_control_2019}. Such
synthesis algorithms typically have strong theoretical guarantees on the correctness,
safety, or stability of the controlled system.

A lot of traditional control theoretical approach to synthesis with STL specifications
have relied on optimizing the quantitative semantics, or \emph{robustness}, of
trajectories generated by the controller.
In~\cite{garg_controllyapunov_2019,lindemann_control_2019}, STL specifications are used
to synthesize control barrier certificates for motion planning, and
in~\cite{sadigh_safe_2016}, the authors develop an extension to STL to reason explicitly
about stochastic systems, named ``Probabilistic STL'' (PrSTL). Quantitative semantics
for STL have been directly encoded into model predictive control optimization problems
to design controllers for robots~\cite{pant_flybylogic_2018,haghighi_control_2019}.

\subsubsection*{Safe Reinforcement Learning}%
\label{sub:tl-rl-related-work}

Temporal logics have been proposed as a means to address various problem in
reinforcement learning. One such problem is \emph{reward hacking}. This refers to cases
where a reinforcement learning agent learns a policy that performs unsafe or unrealistic
actions, though it maximizes the expected total
reward~\cite{leike_ai_2017,amodei_concrete_2016}. A means to approach this problem is
\emph{reward shaping}~\cite{grzes_reward_2017}. The work
in~\cite{aksaray_qlearning_2016} proposes an extension to
Q-learning~\cite{watkins_qlearning_1992} where STL robustness is directly used to define
reward functions over trajectories in an MDP. The approach presented
in~\cite{balakrishnan_structured_2019} translate STL specifications into locally shaped
reward functions using a notion of ``bounded horizon nominal robustness'', while the
authors in~\cite{li_reinforcement_2017} propose a method that augments an MDP with
finite trajectories, and defines reward functions for a truncated form of Linear
Temporal Logic.

Likewise, temporal logics have been used with model-based learning to prove the
safety of a controller that has been designed for the learned models. An example
of this is~\cite{fu_probably_2014}, where the authors propose a PAC learning
algorithm that has been constrained using Linear Temporal Logic specifications.
This is done by learning the transition dynamics of the \emph{product automaton}
between the unknown model and the \emph{omega-regular automaton} corresponding
to the LTL specification, while simultaneously using value-iteration to improve
on the policy. The work by~\cite{berkenkamp_safe_2017} uses Gaussian process
models to learn the system dynamics, while using approximations of
\emph{Lyapunov functions} to design a controller using the learned models. Using
a similar model learning approach, the authors
in~\cite{ohnishi_barriercertified_2019} propose the use of learning a
specialized dynamics model that takes into account safe exploration using
\emph{barrier certificates}, while generating control barrier certificate-based
policies.

\section{Conclusion}%
\label{sec:conclusion}
In this paper, we propose a model-based reinforcement learning technique
that uses simulations of the robotic system to learn a dynamical predictive
model for the system dynamics, represented as a deep neural network (DNN). This 
model is used in a model-predictive control setting to provide finite-horizon 
trajectories of the system behavior for a given sequence of control actions.
We use formal specifications of the  system behavior expressed as Signal 
Temporal Logic (STL) formula, and utilize the quantitative semantics of STL to 
evaluate these finite trajectories. At each time-step an online optimizer 
picks the sequence of actions that yields the lowest cost (best reward) w.r.t. 
the given STL formula, and uses the first action in the sequence as the control 
action. In other words, the DNN model that we learn is used in a receding horizon 
model predictive control scheme with an STL specification defining the 
trajectory cost. The actual optimization over the action space is performed 
using evolutionary strategies. We demonstrate our approach on several case
studies from the robotics and autonomous driving domain.

\bibliographystyle{IEEEtran}
\bibliography{bib}

\end{document}